\RequirePackage{snapshot}
 \documentclass[tablecaption=bottom,wcp]{jmlr} % W&CP article

\usepackage{booktabs}
\usepackage[load-configurations=version-1]{siunitx} % newer version
\theorembodyfont{\upshape}
\theoremheaderfont{\scshape}
\theorempostheader{:}
\theoremsep{\newline}

\jmlrproceedings{}{}

\title[VIB is Half Bayes]{VIB is Half Bayes}

\author{\Name{Alexander A. Alemi} \Email{alemi@google.com}\\
  \Name{Warren R. Morningstar} \Email{wmorning@google.com}\\
  \Name{Ben Poole} \Email{pooleb@google.com}\\
  \Name{Ian Fischer} \Email{iansf@google.com}\\
  \Name{Joshua V. Dillon} \Email{jvdillon@google.com}}
\usepackage{import}
\usepackage{mycustom}

\begin{document}

\maketitle

\begin{abstract}
In discriminative settings
such as regression and classification there are two random variables
at play, the inputs $X$ and the targets $Y$. Here, we demonstrate
that the Variational Information Bottleneck can be viewed as a
compromise between fully empirical and fully Bayesian objectives,
attempting to minimize the risks due to finite sampling of $Y$ only.
We argue that this approach provides some of the benefits of Bayes
while requiring only some of the work.
\end{abstract}

\section{Introduction}
Big models, big data, and maximum likelihood training are a proven recipe for learning powerful and generalizable  neural network models. But training such large models on small data results in overfitting and poor performance. How can we achieve good performance from limited data?

Bayesian inference presents one such mechanism. 
Bayesian inference can be seen as minimizing a PAC-style upper bound on generalization performance from finite data~\citep{pacmbayes,masegosa,germain2016pac}. However, exactly performing Bayesian inference is costly, requiring careful tuning of MCMC methods or expressive variational distributions to match the posterior~\citep{hmcishard,vioverview}.%

Here, we show that training on multiple outputs $Y$ for each input $X$ can be beneficial, and derive a training objective which provides these benefits without actually having to collect multiple outputs for each input. The resulting objective matches the Variational Information Bottleneck (VIB)~\citep{vib}, and provides a tractable alternative to Bayesian inference that loses some of the guarantees but retains much of the qualitative and quantitative performance.

\section{Preliminaries}
Consider training a neural network with parameters $\theta$ to output a stochastic representation $z \sim q(Z|x,\theta)$ for each input $x \in X$. From the representation, we can predict the target $y$ with a fixed (parameter-free) \emph{classifier} or \emph{regressor} $p(y|z)$. At test time, we will form the predictive distribution $q(y|x,\theta) = \int \textrm{d}z\, q(z|x,\theta)p(y|z)$.

Assuming the true data comes from some joint distribution $\nu(X,Y)$, 
we aim to learn a predictive distribution that is as close
as possible to the true conditional distribution $\nu(Y|X)$, as 
measured by the expected conditional KL divergence:\footnote{%
  Since much of the paper is concerned with the differences between taking expectations
  with respect to the true distribution versus the empirical distribution,
  we're using a blue $\Et$ to denote expectations with 
  respect to the true data distribution and red $\Ee$ to
  denote expectations with respect to the empirical distribution
  to increase the visibility of this distinction.
}
\begin{equation}
    \Et_{\nu(X)}\left[ \KL[\nu(Y|X), q(Y|X,\theta)] \right] = \int \textrm{d}x\,\textrm{d}y\, \nu(x,y) \log \frac{\nu(y|x)}{q(y|x,\theta)} \defeq \mathcal P(\theta) - \H[\nu(Y|X)],
\end{equation}
where $\H[\nu(Y|X)]$ is the true conditional entropy and we call $\mathcal P(\theta)$ the \emph{true predictive risk}:
\begin{equation}
    \mathcal P(\theta) \defeq \Et_{\nu(X,Y)} \log \left( -\E_{q(Z|X,\theta)}[ p(Y|Z) ] \right).
    \label{eqn:truepredrisk}
\end{equation}
Invoking Jensen's inequality, we can upper bound $\mathcal P(\theta)$ with the \emph{true classification risk}:
\begin{equation}
    \label{eqn:trueclassrisk}
    \mathcal P(\theta) \leq \mathcal C(\theta) \defeq \Et_{\nu(X,Y)} \E_{q(Z|X,\theta)} [ -\log p(Y|Z) ].
\end{equation}
$\mathcal C(\theta)$ measures how well we can predict the targets given a sample from our representation $z$ of each input $x \in X$.
Unlike $\mathcal P(\theta)$ which contains a $\log$ of an expectation, we can compute unbiased estimates of $\mathcal C$ using Monte-Carlo.

While we don't know the true distribution, we have access to $n$ paired samples from this distribution, a \emph{dataset} $D_n = \{ (x_i, y_i) \}_i^n$. 
We need ways to approximate the true classification
risk (\cref{eqn:trueclassrisk}) while only using a finite number of samples.

Maximum Likelihood (ML) tries to minimize the \emph{empirical classification risk}:
\begin{equation}
    \widehat C(\theta; D_n) \defeq \Ee_{\hat \nu_n(X,Y)} \E_{q(Z|X,\theta)} [ -\log p(Y | Z) ],
    \label{eqn:empclassrisk}
\end{equation}
approximating the expectation with respect to the true distribution
with an average over the observed samples.
From the perspective of variational optimization, ML can concentrate on the deterministic representation 
$q(Z|x,\theta) = \delta( z - f_\theta(x) )$
which best predicts the observed target $y$ for each input $x$. 
Unfortunately, ML with finite samples provides no guaranteed relationship to the true classification risk.
In other words, \cref{eqn:empclassrisk} is neither an upper nor a lower bound on \cref{eqn:trueclassrisk}.

\begin{figure}[htbp]
\floatconts
  {fig:toydeterm}
  {\caption{
  A simple demonstration of (\textit{a}) a neural network overfitting,
  (\textit{b}) its corresponding Bayesian neural network doing much better,
  and (\textit{c}) the neural network trained with multiple target samples also doing very well.
  The true distribution is shown in orange, the network's predictive distribution in blue.
  All three models were fit using the same 10 $x \in X$ samples shown as the blue dots.
  The first two models were trained with a single $y$ for each $x \in X$.
  ML with Multiple Target Samples is trained with many $Y$ samples for each $x \in \mathcal{X}$.
  }}
  {%
    \subfigure[Maximum Likelihood]{\label{fig:determ}%
      \includegraphics[width=0.32\linewidth]{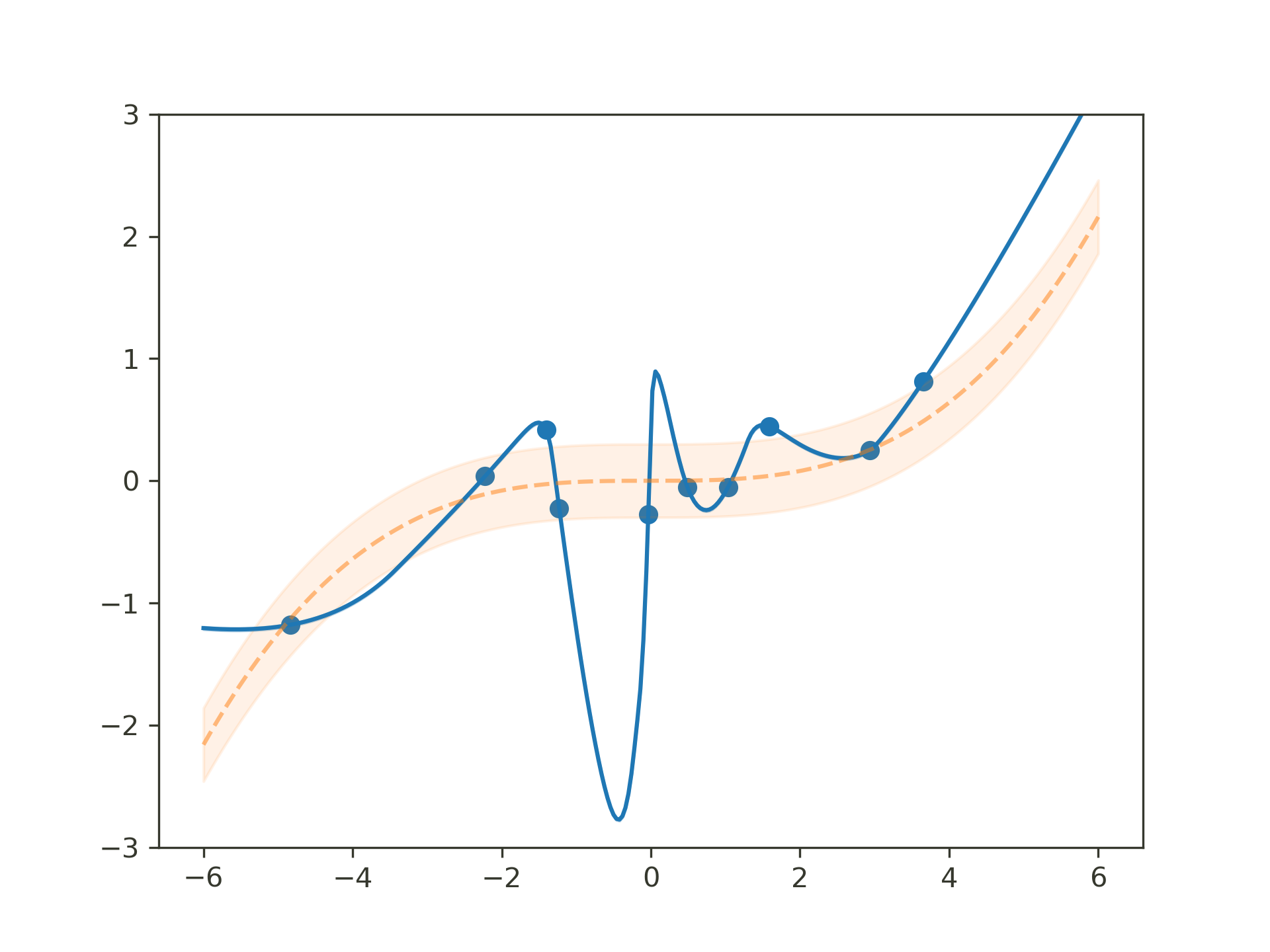}}%
    ~~
    \subfigure[Bayesian neural network (HMC)]{\label{fig:bayes}%
      \includegraphics[width=0.32\linewidth]{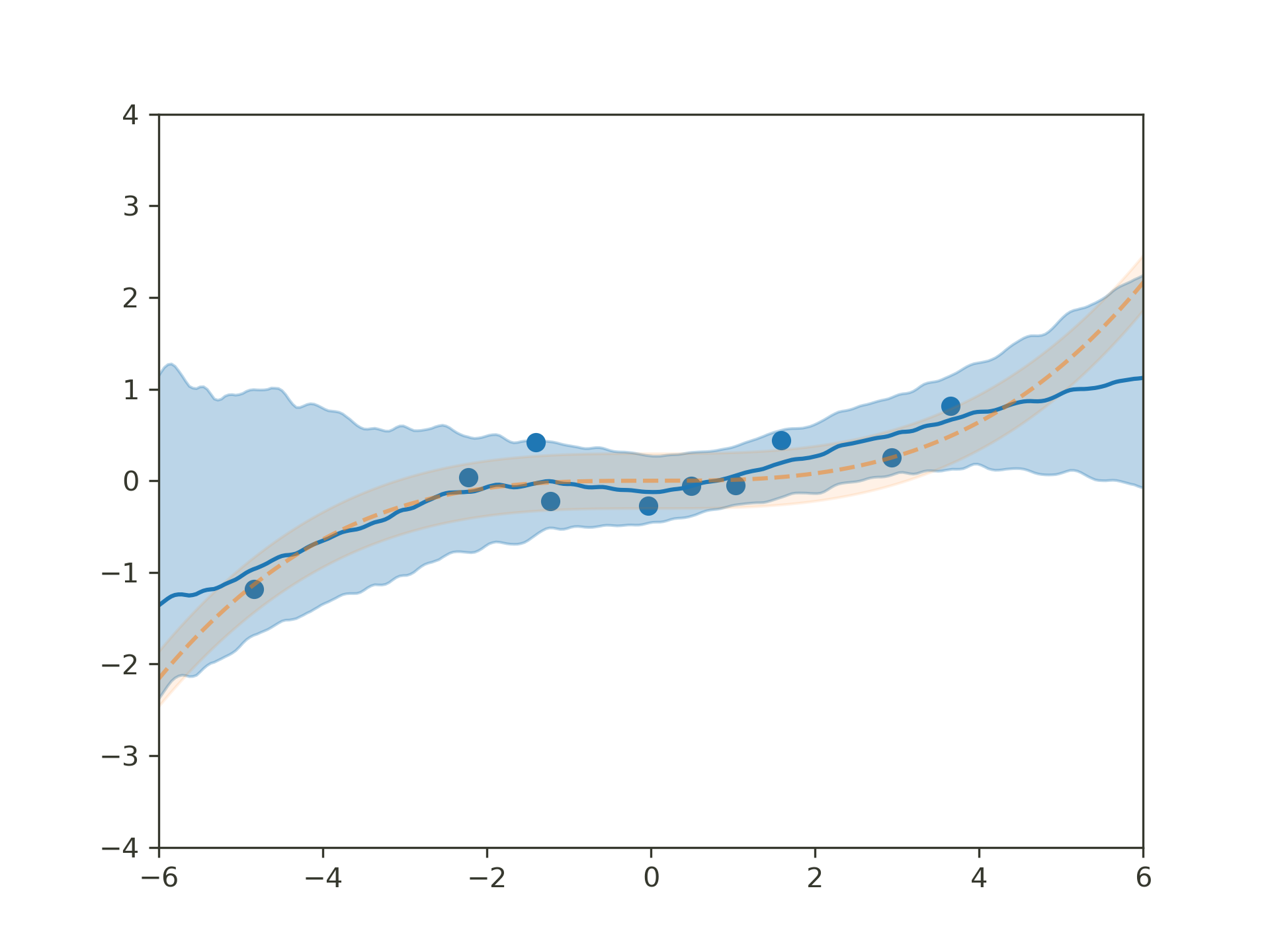}}%
    ~~
    \subfigure[ML with Multiple Target Samples]{\label{fig:multiy}%
      \includegraphics[width=0.32\linewidth]{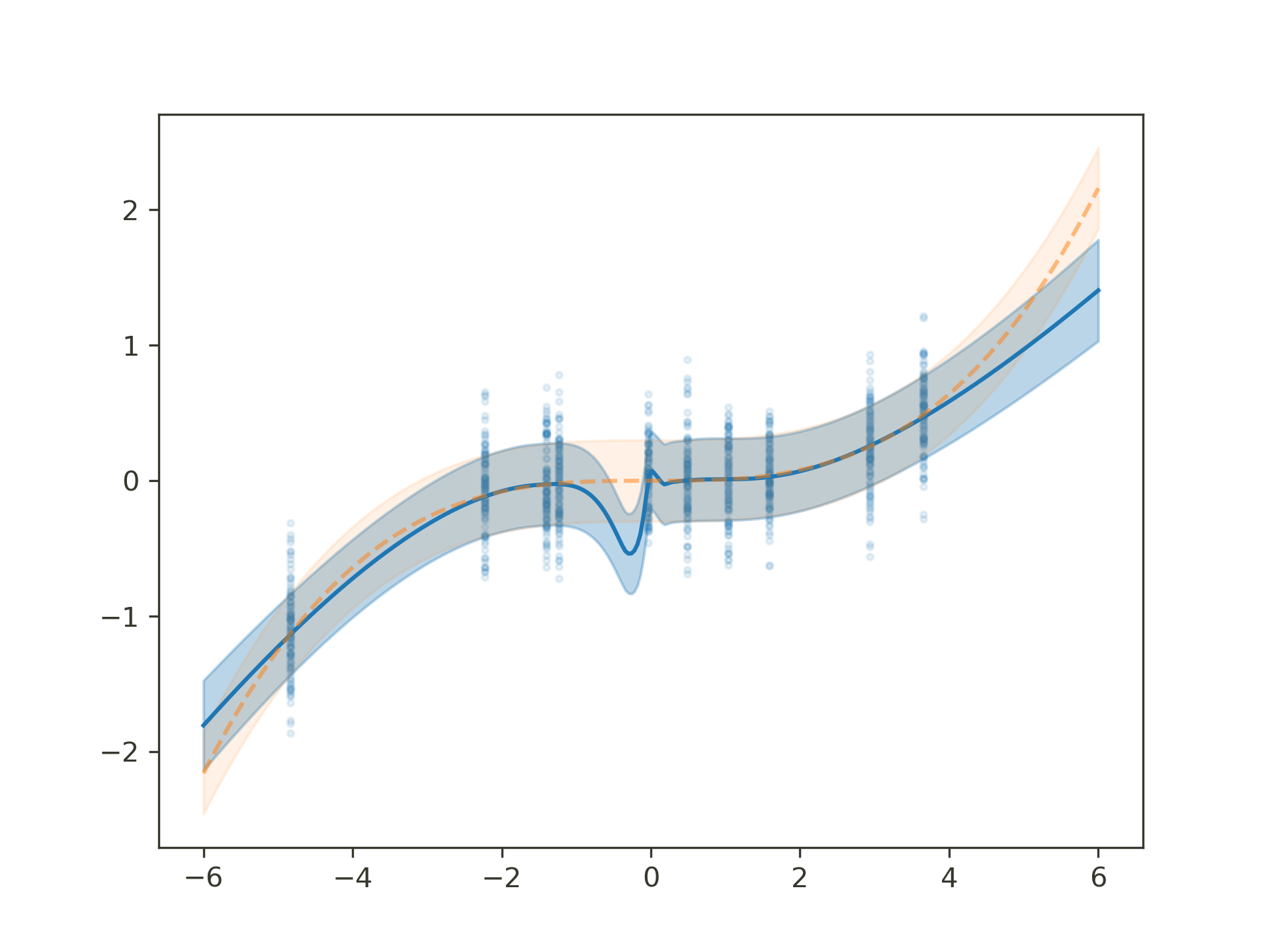}}%
  }
\end{figure}

As an illustration, in \cref{fig:determ} we show what happens if we try to 
minimize \cref{eqn:empclassrisk} 
using a neural network with a two dimensional representation $z = (\mu, \sigma^2)$. This is used to parameterize the mean and standard deviation of a conditional normal distribution $y \sim \Normal(\mu(x), \sigma^2(x))$.
The true model in this case consists of $x$ values uniformly distributed from -5 to 5, and $y$s that are cubic in the $x$s with fixed standard deviation: $y \sim \Normal( x^3/100, 0.3^2)$. 
This true distribution is shown in orange.  The 10 sampled $(x,y)$ pairs
the model was trained on are shown as the blue dots.
The neural network quickly learns to set its predictive
standard deviation to a small value and overfits to the samples.

By approximating the true distribution in \cref{eqn:trueclassrisk} 
with the 
average over the empirical samples, instead of concentrating on the \emph{true} distribution $\nu(X,Y)$, we've had our network attempt to
model the \emph{empirical} distribution $\hat{\nu}_n(x,y) = \frac{1}{n}\sum_i^n \delta(X-x_i)\delta(Y-y_i)$.  The neural network did well at the task 
it was asked to do, but not at the task we wanted.

The traditional way to improve the situation is to try to fit a Bayesian neural network.  Instead of assuming that the parameters of the neural network take on particular values, we treat the parameters $\theta$ as a random variable themselves and compute the posterior over the parameters. %
This noticeably improves the 
predictive distribution, as can be seen in \cref{fig:bayes}, but this 
improvement comes at great computational cost.  In this example,
doing Hamiltonian Monte Carlo to generate samples from the posterior
for the 2178 parameters of the neural network took
8 times longer than training the maximum likelihood model.

\section{Multiple Target Samples}

In the typical setup of a discriminative task, we have a finite sample of pairs $(x,y)$ from the true joint distribution $\nu(X,Y)$.
This amounts to a single sample for the target for each input, a single draw from each 
of the conditional distributions $\nu(Y|x)$.  
Hypothetically, what would happen if we kept the same 10 $X$ samples we used above in \cref{fig:determ,fig:bayes} but collected many $Y$ samples for each? 
In \cref{fig:multiy} we show the result of training precisely the same neural network as in \cref{fig:determ} in this new setup.

With access to many $Y$ samples, the network learns to match the true predictive distribution nearly exactly at those sampled points.  At the 
same time, the neural network does a reasonable job of interpolating
between the sampled points while maintaining a good degree of 
predictive uncertainty.  When asked to extrapolate outside of the data,
the quality of the predictive distribution diminishes noticeably. 
Overall, asking the neural network to match a \emph{half-empirical} distribution $\nu(Y|X) \hat{\nu}_n(X)  = \nu(Y|X) \frac{1}{n}\sum_i^n\delta(X-x_i)$ has produced reasonably good results.
This multiple target setup is similar to problem settings that use \emph{soft targets}, such as teacher-student learning setups~\citep{darkknowledge}, which have proven effective.  Similarly, for image classification tasks, using multiple label samples can lead to improved fits~\citep{cifar10h}.

Is there some way to provide the sorts of guarantees Bayesian inference 
provides, but only with respect to the finite number of $Y$ samples for 
each $X$? Is there some kind of compromise position we could adopt
that achieves performance similar to that in \cref{fig:multiy} without
requiring actually collecting additional target samples for each input?

\section{VIB as PAC-Bayes}

One way to view the source of the Bayesian guarantees is that 
Bayesian inference optimizes a PAC style upper
bound on the \emph{true risk}~\citep{masegosa,pacmbayes}.
By penalizing
the posterior from being too distinct
from the prior,
Bayesian inference \emph{probably} won't overfit (PAC stands for \emph{Probably} Approximately Correct, and the bound that Bayesian inference optimizes holds with high probability even with a finite training sample).

We can invoke the same PAC-Bayes bound as in the Bayesian case,
but only on the inner expectation over \emph{targets} demonstrating that with probability at least $1-\xi_X$ (see \cref{chap:proofs} for proofs):
\begin{multline}
    \mathcal C(\theta) \leq \Et_{\nu(X)}\Bigg[ \E_{q(Z|X,\theta)}\left[\Ee_{\hat{\nu}_n(Y|X)} [-\log p(Y|Z)] \right] + \tau \KL\left[ q(Z|X,\theta), r(Z) \right] \\
    + \tau \log \E_{r(Z)}\Et_{\nu(Y|X)}\left[ e^{\frac 1 \tau \left( \left( \Et_{\nu(Y|X)}[-\log p(Y|Z)] \right) - \left( \Ee_{\hat{\nu}_n(Y|X)}[ -\log p(Y|Z)] \right) \right) } \right] - \log \xi_X \Bigg].
    \label{eqn:vibpac}
\end{multline}
While this is a lot to unpack, notice that all of the terms on the second line are constants with respect to the neural network's representation $q(Z|X,\theta)$, and so can be dropped in its objective.  
\Cref{eqn:vibpac} (nearly always) provides an upper bound on the \emph{true classification risk},
However, it is still intractable as it includes an expectation over $\nu(X)$.

With this observation, we could instead adopt a mixed approach.
Why not take the Bayesian strategy of minimizing an upper bound with respect to the conditional expectation 
of targets $\nu(Y|X)$ while using the bold Maximum Likelihood strategy of a drop-in Monte Carlo estimate for the expectation over inputs $X$?
Doing so gives us:
\begin{equation}
    {\sf VIB}(\theta) \defeq \Ee_{\hat{\nu}_n(X)} \E_{q(Z|X,\theta)}\left[ \Ee_{\hat{\nu}_n(Y|X)} \left[ -\log p(Y|Z) \right] + \tau \log \frac{q(Z|X,\theta)}{r(Z)} \right].
    \label{eqn:vib}
\end{equation}
This objective is equivalent to the Variational Information Bottleneck (VIB) objective of~\citet{vib}.  
The VIB objective was originally motivated as
being a variatonal upper bound on the
Information Bottleneck objective~\citep{ib}:
\begin{equation}
    \max I(Z;Y) - \tau I(Z;X).
\end{equation}
$I(X;Y) \defeq \Et_{p(X,Y)}\left[ \log \frac{p(X,Y)}{p(X)p(Y)} \right]$ is the \emph{mutual information} between $X$ and $Y$.
The Information Bottleneck aims to find a representation $Z$ that is as maximally informative about the target $Y$ as possible ($I(Z;Y)$), subject to a constraint
on how \emph{expensive} that representation is, measured by
how many bits about the input it retains ($I(Z;X)$).

Here we have stumbled upon an alternative motivation of the same objective, showing that the VIB objective can be seen as \emph{half Bayesian}.
VIB attempts to protect against overfitting on a finite number of sampled targets for each input without addressing potentially overfitting to the finite number of sampled inputs themselves.
It tries to concentrate on the half-empirical distribution of \cref{fig:multiy}.
The VIB objective does not itself provide any bound on the true classification risk, just as Maximum Likelihood does not.
Yet, VIB style objectives have been shown to improve model's generalization and robustness~\citep{cebrobust}.

Where building a traditional Bayesian neural network requires a distribution over all of the parameters of the network, solving \cref{eqn:vib} only requires a distribution over the \emph{output activations} of the network.
This is a much lower dimensional space and much easier to deal with computationally.
In the VIB setup, the output of the neural network is made an explicit distribution on the representation space, often chosen to be a Gaussian distribution for simplicity.

Notice that in this interpretation, we are not allowed to learn either the 
\emph{classifier} distribution $p(Y|Z)$ or the \emph{prior} or \emph{marginal} $r(Z)$
using \cref{eqn:vibpac}, as both of those distributions appear in
the second line but are dropped in the objective~(\cref{eqn:vib}). In this way
this half-Bayesian interpretation differs from the existing VIB
literature, where both $p(Y|Z)$ and $r(Z)$ are thought to be 
variational approximations that are free to be fit simultaneously
with the representation $q(Z|X,\theta)$.  If the data were split, or there were 
additional holdout data, these could be used to refine either $p(Y|Z)$ or
$r(Z)$ similar to the setup in~\citet{datapac}. 

If we wanted to generate a fully valid bound on the true classification risk, 
we could continue the road we are on and simple apply another PAC-Bound to \cref{eqn:vibpac}, this time with respect to the \emph{parameters} of the encoding distribution $q(Z|X,\theta)$.  See \cref{chap:proofs} for the full details, but dropping the constant terms with regards to the 
objective we obtain a fully 
\emph{Bayesian variational information bottleneck}:
\begin{equation}
    {\sf BVIB}[q(\Theta)] \defeq \Ee_{\hat{\nu}_n(X)} \E_{q(\Theta)} \E_{q(Z|X,\Theta)} \left[ \Ee_{\hat{\nu}_n(Y|X)}\left[ -\log p(Y|Z) \right]  + \tau \log \frac{q(Z|X,\Theta)}{r(Z)} + \frac{\gamma}{n} \log \frac{q(\Theta)}{r(\Theta)} \right]
    \label{eqn:bvib}
\end{equation}
Realizing \cref{eqn:bvib} 
could be as simple as adding weight decay to the parameters of the 
representation in \cref{eqn:vib}.
Objectives of this sort also appeared in~\citet{therml}, where again they
were motivated from an alternative, information theoretical perspective.
\section{Demonstration}

\begin{figure}[htbp]
\floatconts
  {fig:vib}
  {\caption{A simple demonstration that VIB can learn 
  to capture uncertainty in much the same way that we could if we
  trained with multiple target samples as in \cref{fig:multiy}.
  The three figures show different values for $\tau$.
  All three models have the same failure as seen in \cref{fig:determ} and \cref{fig:multiy} because they were all initialized with the same random seed.
  This highlights the inherent risk of training with the empirical sample -- the model can make arbitrary errors away from the observed data.
  }}
  {%
    \subfigure[VIB $\tau=10^3$]{\label{fig:viblow}%
      \includegraphics[width=0.32\linewidth]{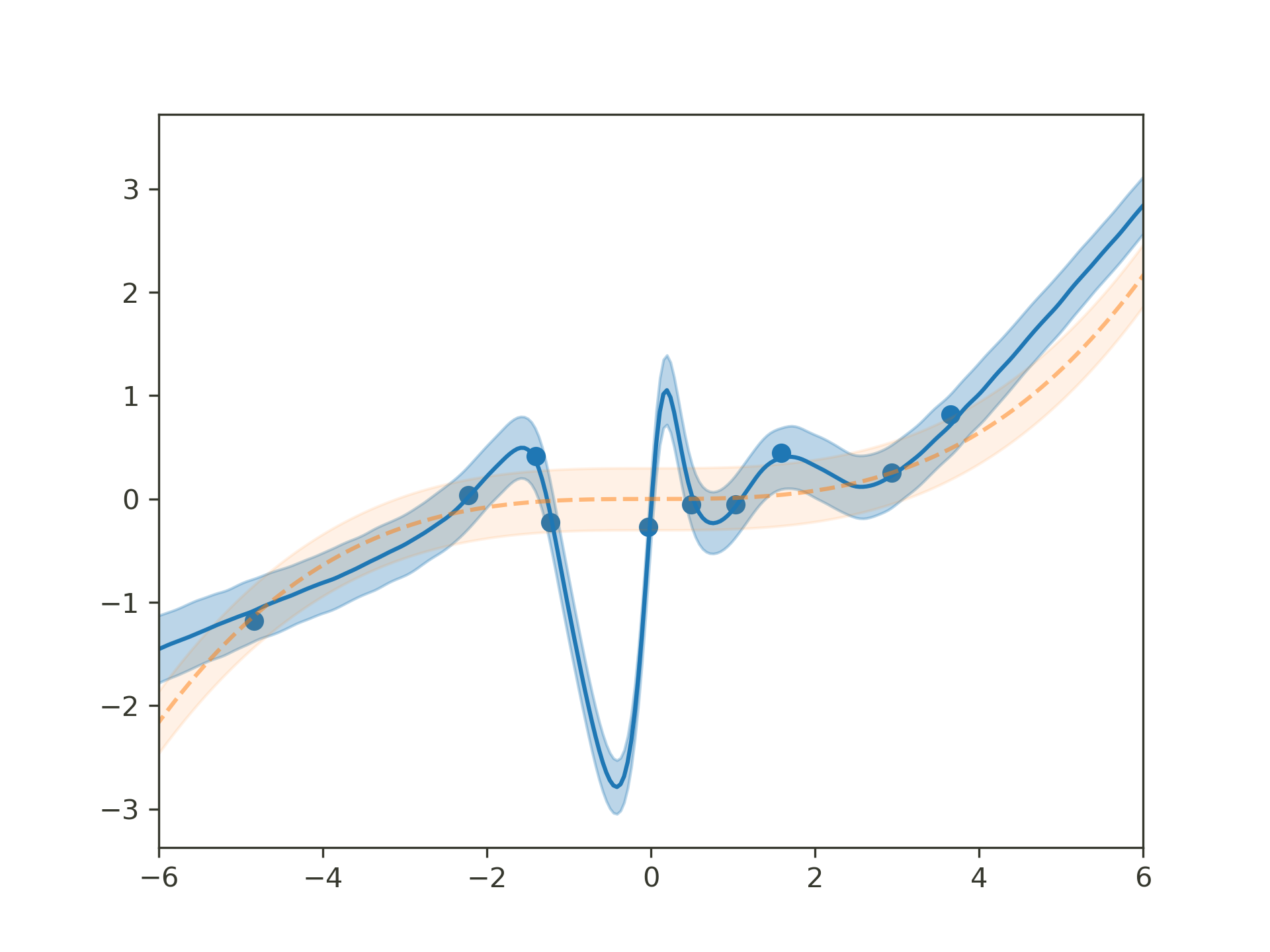}}%
    \subfigure[VIB $\tau=10^4$]{\label{fig:vibmid}%
      \includegraphics[width=0.32\linewidth]{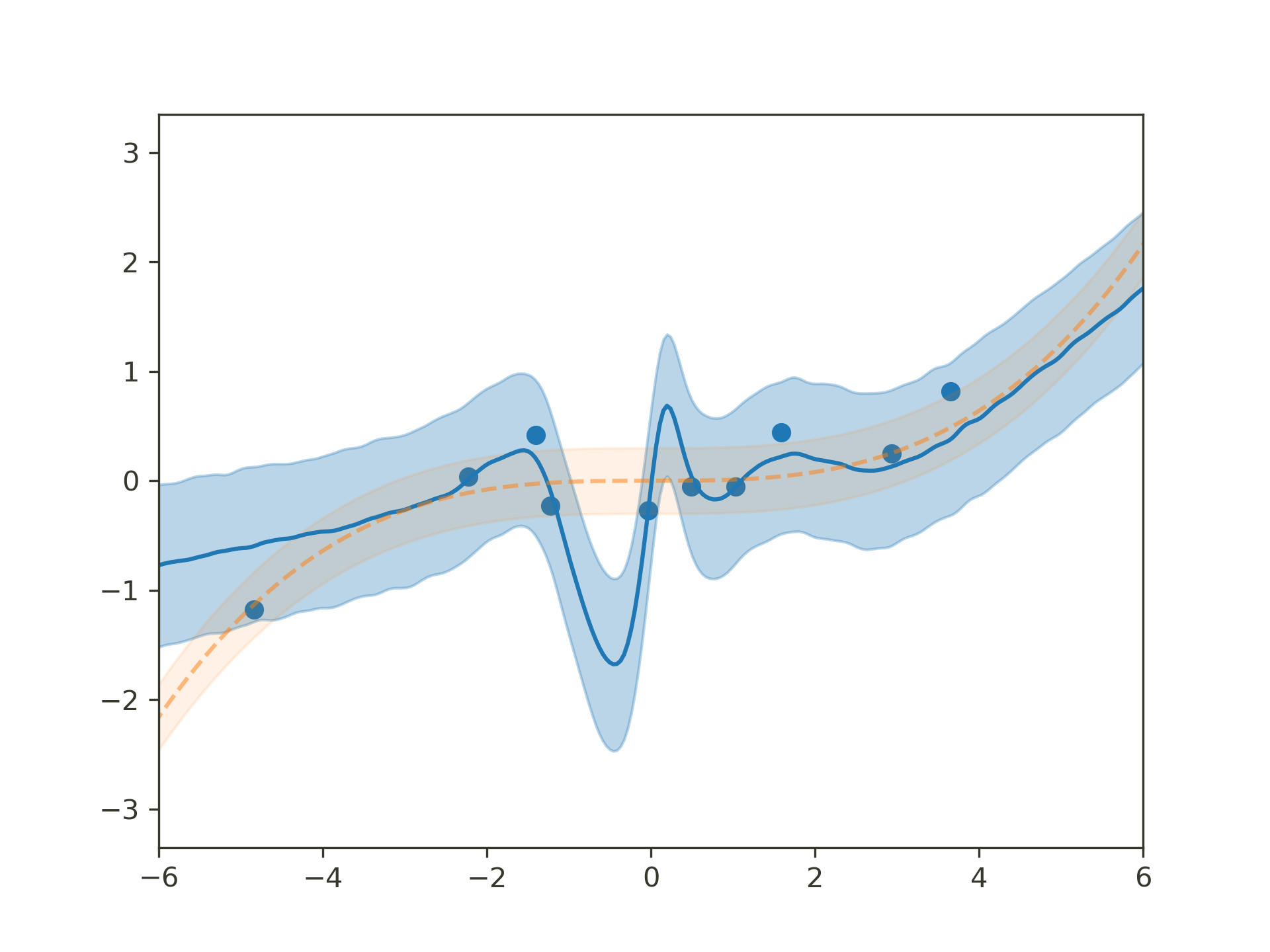}}
    \subfigure[VIB $\tau=10^5$]{\label{fig:vibhigh}%
      \includegraphics[width=0.32\linewidth]{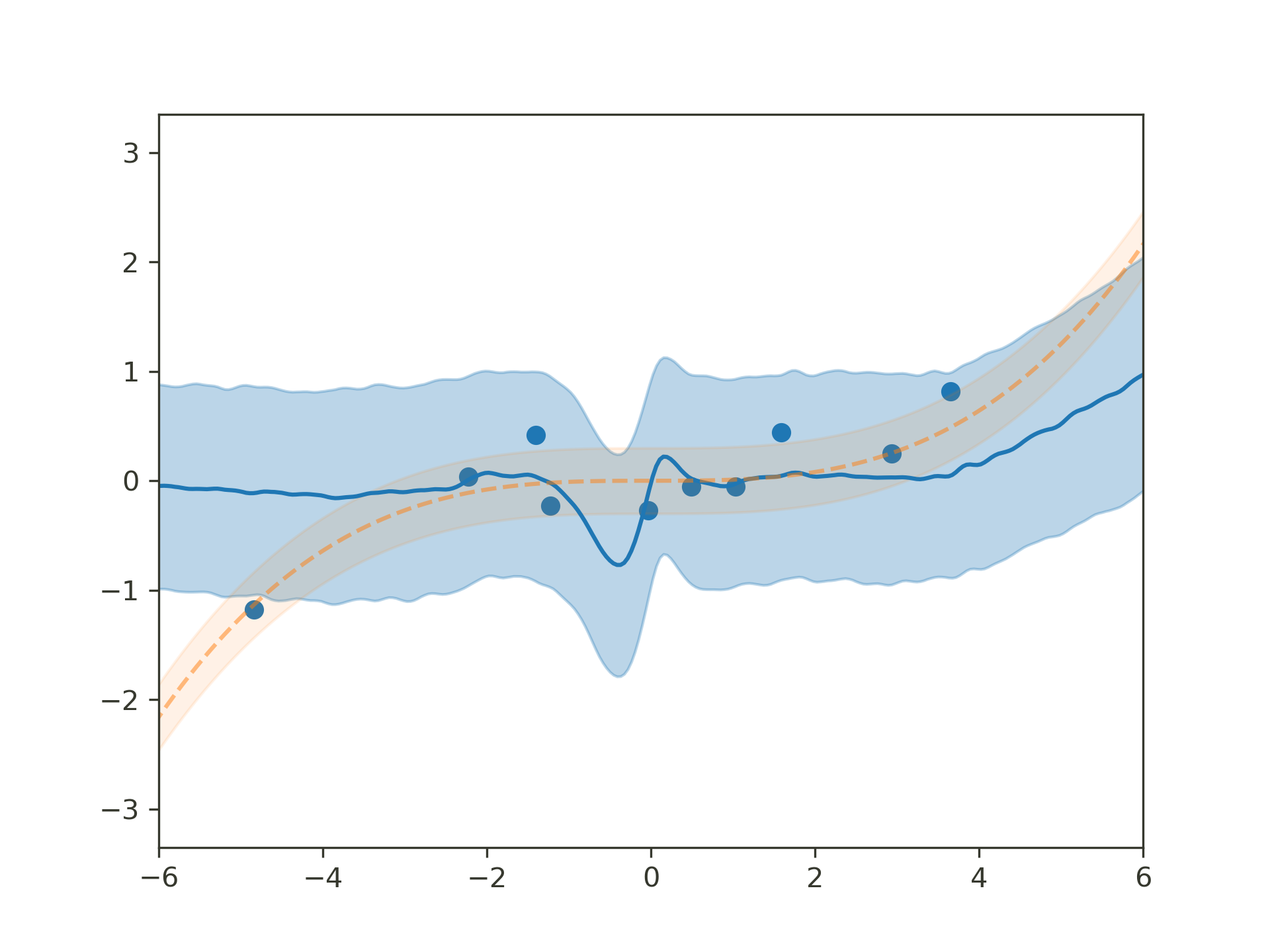}}
  }
\end{figure}

To illustrate that this can work, in \cref{fig:vib} we show the result
of fitting the VIB objective (\cref{eqn:vib}) to the same 10 data points as in \cref{fig:toydeterm,fig:multiy}, 
using the same random network initialization.  
The results are sensitive to the choice of $\tau$, so we show
several values near the best performing models.
Full experimental details can be found in \cref{sec:details}.
\Cref{fig:vibmid} in particular
has a similar predictive distribution
to \cref{fig:multiy}, while only having access to a single 
target sample for each of the 10 input samples shown.

This qualitative sense that the VIB methods are doing well can
be verified quantitatively. In \cref{tab:toy} we show the computed
KL divergences between the true conditional distribution $\nu(Y|X)$ and the predictive distributions $q(Y|X,\theta) = \int \textrm{d}z\, q(z|X,\theta) p(Y|z)$ for each method.  This conditional KL can then be computed in expectation both 
with respect to the empirical $X$ distribution $\hat{\nu}(X)$ (simply the average on the 10 samples), or in expectation with respect to the true $\nu(X)$, marginalizing from $x=-5$ to $x=5$ uniformly.  
This assesses how well the methods did at learning the predictive distribution both on the values
they were given ($\overline{\KL}$) as well as on all values ($\KL$). The VIB
approaches are competitive with the fully Bayesian model, while
being significantly cheaper to optimize. The VIB 
models did not take noticeably longer to train than the ML model. We give additional experimental results on MNIST \emph{classification} in \cref{sec:mnist}.

\begin{table}[htbp]
    \centering
    \small
    \begin{tabular}{r|c|c|c|c|c|c}
  & Determ & MultiY & Bayes & VIB $10^3$ & VIB $10^4$ & VIB $10^5$ \\
  \hline 
 $\KL$            & 3850 & 0.0993              & 0.195 & 1130 & 1.08  & 1.38  \\
 $\overline{\KL}$ & 1090 & $4.39\times10^{-4}$ & 0.330 & 1.85 & 0.763 & 1.22          
    \end{tabular}
    \caption{True and Empirical KL divergences for the predictive distribution from each method on the toy problem.
    All KLs are measured in bits.
    The large value for $\overline{\KL}$ for Determ is due to the fact that we know the true $\sigma^2$ for $\nu$ -- even interpolating the sampled points doesn't protect against a large empirical risk.
    }
    \label{tab:toy}
\end{table}

\section{Conclusion}

We've demonstrated that on a simple problem 
we can provide most of the benefits of
Bayesian inference for signficantly less work.  
The Variational Information Bottleneck method of~\citet{vib} can be thought of as
a half-Bayesian approach that offers some assurance that it won't too
severely overfit, but only with regards to the finite sampling
of the targets in a discriminative modeling task.

%\acks{Acknowledgements go here.}

\clearpage
\bibliography{bib}

\appendix
%\clearpage

%\import{tex/}{oldBayesPAC}

\section{Theory}
\label{chap:proofs}

In this appendix we prove the claims in the paper.

Suppose $(X,Y)^n\iid \nu(Y|X)\nu(X);$ write their empirical distributions as,
\begin{align}
\hat{\nu}_n(Y,X) &= \frac{1}{n}\sum_i^n \delta(y_i-Y)\delta(x_i-X) \\
\hat{\nu}_n(X) &= \int_\mathcal{Y} \textrm{d} \mu(y)\, \hat{\nu}_n(y,X) = \frac{1}{n} \sum_i^n \delta(x_i-X) \\
\hat{\nu}_n(Y|X) &= \frac{\hat{\nu}_n(Y,X)}{\hat{\nu}_n(X)} = \frac{\sum_i^n \delta(y_i-Y)\delta(x_i-X)}{\sum_i^n \delta(x_i-X)} \label{eq:conditional_empirical}.
\end{align}
For notational simplicity, we regard $\hat{\nu}_n(Y|X)$ as $0$ (for all $y$) for $X\not\in\{x_i\}_i^n.$ Depending on its context, the symbol $\delta$ denotes either the Dirac or Kronecker delta function. (These two caveats are our only notational abuses in the paper.) 

\begin{theorem}\label{thm:double_pac_bayes}
For all $q(\Theta)$ absolutely continuous with respect to $r(\Theta)$, 
$q(Z|X,\Theta)$ absolutely continuous with respect to $r(Z)$
  for all $\{\Theta \in \mathcal{T}:q(\Theta)>0\}$
  and $\{X \in \mathcal{X}:\nu(X)>0\}$,
$(X,Y)^n\iid \nu(Y|X)\nu(X)$,
$\beta_X,\beta\in\mathbb{R}_+$, $n\in \mathbb{N}$, and
$\xi_X,\xi \in (0,1]$,
then with probability at least $1-\max(\xi_X,\xi)$:
\begin{align}
- &\E_{\nu(X)\nu(Y|X)} \log \E_{q(\Theta)q(Z|X,\Theta)}\left[ p(Y|Z) \right] \\
&\le -\frac{1}{n}\sum_i^n \E_{q(\Theta)q(Z|x_i,\Theta)} \left[ \log  p(y_i|Z) \right] \label{eq:expected_distortion} \\
&\qquad + \frac{1}{\beta_X}  \frac{1}{n}\sum_i^n \E_{q(\Theta)} \KL\left[q(Z|x_i,\Theta), r(Z)\right] \label{eq:expected_rate} \\
&\qquad + \frac{1}{\beta} \frac{1}{n} \KL\left[q(\Theta), r(\Theta)\right] \label{eq:encoder_regularizer} \\
&\qquad + \E_{\nu(X)}\left[ \psi_X\left(\nu(Y|X),r(Z),p(Y|Z), \beta_X, \xi_X\right) \right]\\
&\qquad + \psi\left(\nu(X)\nu(\hat{\nu}_n(Y|X)|X), r(\Theta)r(Z),q(Z|X,\Theta)p(Y|Z), n, \beta_X,\beta,\xi\right)
\end{align}

where,

\begin{align*}
\psi_X(\nu(Y|X),&r(Z),p(Y|Z), \beta_X, \xi_X) = \\
&= \frac{1}{\beta_X} \log \E_{r(Z)} \E_{\nu(Y|X)} \left[ e^{ \beta_X \Delta_X }\right] - \frac{\log \xi_X}{\beta_X}\\
\psi(\nu(X) &\nu(\hat{\nu}_n(Y|X)), r(\Theta)r(Z),q(Z|X,\Theta)p(Y|Z), n, \beta_X,\beta,\xi) = \\
&= \frac{1}{\beta n} \log \E_{r(\Theta)} \E_{\nu(X)} \E_{\nu(\hat{\nu}_n(Y|X))} \left[ e^{ \beta n\Delta } \right] - \frac{\log \xi}{\beta n}
\end{align*}

and where $\Delta_X,\Delta$ are defined by Equations \ref{eq:delta_x} and \ref{eq:delta}
and $\nu(\hat{\nu}_n(Y|X))$ is the true probability of the empirical conditional measure.

Note that neither $\psi_X$ nor $\psi$ are a function of $q(\Theta)$ and that quantities \ref{eq:expected_distortion}, \ref{eq:expected_rate}, and \ref{eq:encoder_regularizer} are not a function of the unknowable true data generating distribution, $\nu(Y|X)\nu(X).$

\begin{proof}

First, with probability at least $1-\xi_X$ we have:

\begin{align}
- &\E_{\nu(X)\nu(Y|X)} \log \E_{q(\Theta)} \E_{q(Z|X,\Theta)}\left[ p(Y|Z) \right] \\
&\le - \E_{q(\Theta)} \E_{\nu(X)}  \E_{q(Z|X,\Theta)} \E_{\nu(Y|X)}    \log \left[ p(Y|Z) \right] \label{eq:thm1_part1_jenson}\\
&\underset{\mathscale{0.4}{\xi_X}}{\lesssim} \E_{q(\Theta)} \E_{\nu(X)}  \Bigg[ -\E_{q(Z|X,\Theta)} \E_{\hat{\nu}_n(Y|X)} \log \left[ p(Y|Z) \right] \nonumber\\
&\qquad\qquad\qquad\qquad + \frac{1}{\beta_X}\KL\left[q(Z|X,\Theta), r(Z)\right] \nonumber\\
&\qquad\qquad\qquad\qquad + \psi_X\left(\nu(Y|X), \beta_X, r(Z),\xi_X\right) \Bigg] \label{eq:thm1_part1_compress_logmarkov}
\end{align}

Inequality~\ref{eq:thm1_part1_jenson} follows from Jensen's inequality and inequality \ref{eq:thm1_part1_compress_logmarkov} holds with probability at least $1-\xi_X$ and follows from applying Lemmas \ref{lma:compression} and \ref{lma:logmarkov} to:

\begin{equation}
\Delta_X = \E_{\nu(Y|X)}\left[ -\log p(Y|Z) \right] - \E_{\hat{\nu}_n(Y|X)}\left[-\log p(Y|X)\right].\label{eq:delta_x}
\end{equation}

Again applying Lemmas \ref{lma:compression} and \ref{lma:logmarkov} to:
\begin{align}
\Delta &=   \E_{\nu(X) \hat{\nu}_n(Y|X)}  \left[
          - \E_{q(Z|X,\Theta)}  \log \left[ p(Y|Z) \right] 
          + \frac{1}{\beta_X} \KL\left[q(Z|X,\Theta), r(Z)\right]
       \right] - \nonumber \\
   &\qquad
       \E_{\hat{\nu}_n(X)\hat{\nu}_n(Y|X)} \left[
          - \E_{q(Z|X,\Theta)}  \log \left[ p(Y|Z) \right] 
          + \frac{1}{\beta_X } \KL\left[q(Z|X,\Theta), r(Z)\right]
       \right]\label{eq:delta}.
\end{align}

we conclude that with probability at least $1-\max(\xi_X,\xi)$:
\begin{align*}
\text{[Eq} & \text{uation \ref{eq:thm1_part1_compress_logmarkov}]} \\
&\underset{\mathscale{0.4}{\xi_X,\xi}}{\lesssim} -\E_{q(\Theta)} \E_{\hat{\nu}_n(X,Y)} \E_{q(Z|X,\Theta)} \log \left[ p(Y|Z) \right] 
 \nonumber\\
&\qquad + \frac{1}{\beta_X} \E_{q(\Theta)}\E_{\hat{\nu}_n(X)} \left[ \KL\left[q(Z|X,\Theta), r(Z)\right] \right] \\
&\qquad + \frac{1}{\beta n} \KL\left[q(\Theta), r(\Theta)\right] \\
&\qquad + \E_{\nu(X)}\left[ \psi_X\left(\nu(Y|X),r(Z),p(Y|Z), \beta_X, \xi_X\right) \right]\\
&\qquad +  \psi\left(\nu(X)\nu(\hat{\nu}_n(Y|X)), r(\Theta)r(Z),q(Z|X,\Theta)p(Y|Z), n, \beta_X,\beta,\xi\right).
\end{align*}
The proof is completed by expanding occurrences of $\E_{\hat{\nu}_n}$ as a summation.
\end{proof}
\end{theorem}

The proof of Theorem~\ref{thm:double_pac_bayes} is similar to twice applying the technique of \cite{pacmbayes} (with $m=1$). The \cite{pacmbayes} proof followed arguments similar to \cite{masegosa} which itself followed arguments similar to \cite{germain2016pac}.

Note that in the text body we used $\tau=\frac{1}{\beta_X}$ and $\gamma=\frac{1}{\beta}$. 

We also note that one can use Lemma \ref{lma:conditional_empirical} to rewrite $\E_{\hat{\nu}_n}$ expectations as conditional averages (e.g., this could be done to  \cref{eqn:vibpac}).

\subsection{Lemmas}

In this section we present several Lemmas used to simplify this paper's proofs. The Lemmas are well-known and are given here for the reader's convenience. 

\begin{lemma}[Compression]\label{lma:compression}
If $p(\Theta)$ is absolutely semicontinuous wrt $r(\Theta)$  and
$\E_{r(\Theta)}[e^{f(\Theta)}] < \infty$, then
$\E_{p(\Theta)}[f(\Theta)] \le \KL\left[p(\Theta), r(\Theta)\right] + \log \E_{r(\Theta)}[e^{f(\Theta)}].$
\begin{proof}
Write $q(\Theta) \defeq \frac{r(\Theta)e^{f(\Theta)}}{\E_{r(\Theta)} [
e^{f(\Theta)}]}$ and note that Lemma~\ref{lma:gibbs} implies,
$0 \le \KL\left[p(\Theta), q(\Theta)\right]
= \KL\left[p(\Theta), r(\Theta)\right] - \E_{p(\Theta)} [ f(\Theta) ] + \log \E_{r(\Theta)}[ e^{f(\Theta)} ].$
\end{proof}
\end{lemma}
Proof due to \cite{banerjee2006bayesian,zhang2006information}.

\begin{lemma}[Log Markov Inequality]\label{lma:logmarkov}
For any $\xi \in (0,1]$ and random variable $Z \sim p$ with $p(Z \le 0) = 0$ then $p(\log Z \le \log \E_p[Z] - \log \xi) \ge 1-\xi.$
\begin{proof}
Markov's inequality states that $p(Z > t) \le \frac{\E_p[Z]}{t}$ for non-negative random variable $Z \sim p$ and $t>0$. Substituting $t=\frac{\E_p[Z]}{\xi}$ implies
$p(Z > \frac{\E_p[Z]}{\xi}) \le \xi$. Combining this with the fact that
$\log$ is a non-decreasing bijection implies $p(\log Z > \log \E_p[Z] - \log
\xi)\le \xi.$ Examining the complement interval completes the proof.
\end{proof}
\end{lemma}

\begin{lemma}[Gibb's Inequality]\label{lma:gibbs}
If $p(\Theta)$ is absolutely semicontinuous wrt $r(\Theta)$, then $\KL[p, q] \ge 0.$
\begin{proof}
$\KL[p,q] = -\E_{p(x)}\left[ \log \frac{q(x)}{p(x)} \right] \ge -\log \E_{p(x)}\left[\frac{q(x)}{p(x)}\right]=-\log 1=0$
where the inequality is Jensen's.
\end{proof}
\end{lemma}

\begin{lemma}[Conditional Empirical Expectation]\label{lma:conditional_empirical}
Assuming $X \in \{x_i\}_i^n,$ then:
\begin{equation}
\E_{\hat{\nu}_n(Y|X)}[ f(Y,X) ] = \frac{1}{\sum_i^n\delta(x_i-X)} \sum_i^n \delta(x_i-X)f(y_i,x_i).
\end{equation}
\begin{proof}
$\E_{\hat{\nu}_n(Y|X)}[ f(Y,X) ]
=\int_\mathcal{Y} \textrm{d}\mu(y)\, \hat{\nu}_n(y|X) f(y,X) 
= \int_\mathcal{Y} \textrm{d}\mu(y)\, \sum_i^n \frac{\delta(y_i-y)\delta(x_i-X)}{\sum_j^n \delta(x_j-X)}   f(y,X) =  \sum_i^n \delta(x_i-X)\frac{\int_\mathcal{Y} \textrm{d}\mu(y)\,\delta(y_i-y)f(y,X)}{\sum_j^n \delta(x_j-X)} = \frac{1}{\sum_i^n \delta(x_i-X)} \sum_i^n \delta(x_i-X)f(y_i,x_i).$
The $\delta$ function is either the Dirac or Kronecker delta function, depending on whether measure $\mu$ is continuous or discrete.
\end{proof}
\end{lemma}

\section{Experimental Details}
\label{sec:details}

For~\cref{fig:determ,fig:bayes,fig:multiy,fig:viblow,fig:vibmid,fig:vibhigh}
all experiments were done with \textsf{JAX}~\citep{jax}.

The true data distribution was taken to be
\begin{align*}
    X &\sim {\sf Uniform}(-5, 5) \\
    Y|X &\sim {\sf Normal}\left( \frac{x^3}{100}, 0.3^2 \right).
\end{align*}

Ten samples were taken for the data distribution.  The predictive network
consisted of two fully connected layers with 32 hidden units followed by an
\textsf{elu} activation.  The final layer was a linear layer with 2 outputs,
the first of which was taken as the mean, and the
second generated the standard deviation of the predictive model
with a softplus activation and a minimum value of 0.01:  ($\sigma^2 = 0.01 + {\sf softplus}(x)$).

A standard Lecun style truncated normal initialization scheme 
was used for the kernels, and the biases were initialized to be zero.  The initial parameter variance was increased by a factor of 5, 
which was found to be 
important to get the one dimensional networks to converge well on the 
range and domain of the toy problem.  All problems used the same
initial parameters and the same \textsf{adabelief} optimizer
with a \textsf{cosine} decay schedule on the learning rate
starting at $10^{-3}$ and ending at 0 after 100k steps, the length of 
each optimization run.

To sample from the Bayesian neural network, \textsf{tensorflow\_probability}'s \textsf{JAX} backend Hamiltonian Monte Carlo sampler was used \citep{dillon2017tfp,lao2020tfpmcmc}. In particular 1000 results were generated from the chain with 10k burn-in steps, dual averaging
step size adaptation with a step size of $10^{-3}$ and 100 leapfrog steps, 9000 adaptation steps and a target acceptance probability of 0.7.
The initialization distribution used for the neural network experiments was
taken to be the prior distribution, both for the kernel and bias parameters.

For the VIB experiments, the same neural network as above was used to form the representation $q(Z|X)$. The classifier network $p(Y|Z)$ was taken to be
a Normal distribution with a small fixed variance: ${\sf Normal}(z, 0.01^2)$.  The marginal $r(Z)$ was set to be a fixed unit Normal: ${\sf Normal}(0,1)$.

All KL divergences were computed using 100k samples of data points and 10k samples from the any intermediate distributions as required.

\section{MNIST Experiments}
\label{sec:mnist}

In the main paper, we demonstrate that VIB learns a model which retains uncertainty about the targets $Y$, even if it provides no guarantees about generalization in $X$.  For an additional illustrative comparison, we train models on the MNIST dataset.  Specifically, we compare the predictive models of a deterministic deep neural network and a VIB model having the same architecture.  We use a parameter free decoder, with a categorical likelihood, multivariate normal prior, and multivariate normal posterior.  For the prior, we assume zero mean and identity covariance, while we predict the full covariance matrix for the posterior.  The deterministic model replicates this setup, but simply predicts the categorical likelihood rather than the posterior, and it has no prior.  We train models for 50 epochs with a learning rate of 0.001, which is decayed by half every 5000 steps.  We further use a batch size of 128.  For the VIB model, we evaluate the objective using 4 samples from the posterior and use $\tau=0.005$ as a weighting for the $\KL$ penalty.  

\begin{figure}[!htb]
    \centering
    \includegraphics[width=\hsize]{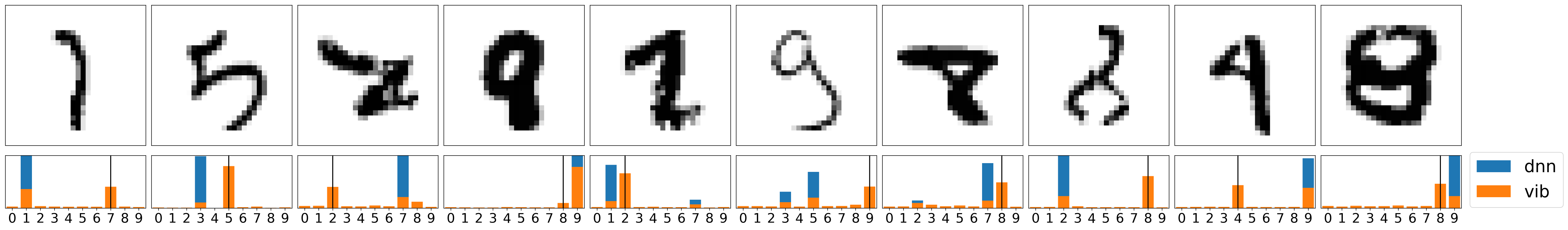}
    \caption{A selection of images from the MNIST dataset which are misclassified, either by a deterministic model or by a VIB model.  The lower panel shows the class probabilities predicted by the model for a VIB and a deterministic model.  We find that the deterministic model tends to assign high probability to a single class, while a VIB model tends to incorporate uncertainty between multiple classes.}
    \label{fig:mnist}
\end{figure}

We evaluate models using the classification accuracy as well as the log-likelihood of the test set.  To compute the log-likelihood of VIB models, we marginalize over 1000 samples from the posterior.  Both models produce similar final test set accuracies (99.2\% for the deterministic model, and 99.4\% for the VIB model).  We further find that the test set log-likelihood for the VIB model is higher (-508 versus -786 for the deterministic model).  However, this later finding is heavily dependent on the $\tau$ used in training:  We find that the VIB model can measure a lower log-likelihood if the $\tau$ multiplier is larger.  Note that the accuracy is more robust to this hyperparameter, and we find that the VIB model consistently observes higher accuracy than the deterministic model over the range of $\tau$ we explored.

In addition to showing that VIB leads to models which have a higher test log-likelihood and accuracy, we also examined if the resulting predictive models learned by VIB incorporate more uncertainty into the labels than do those which are trained with a deterministic network.  For this, in \cref{fig:mnist} we show 10 images from the test set, each of which is classified incorrectly by either the VIB model or by the deterministic model.  Below each image, we show the class probabilities predicted by the deterministic and VIB models, and indicate the true class with a vertical line.  Our main observation is that many of the deterministic models overpredict the probability of a label, with 7 out of 10 images being assigned a class probability greater than 95\%.  We also find that in these same situations, VIB often folds additional probability into other classes one of which is typically the correct class.  This, when combined with the higher test set log-likelihood agrees with our findings that VIB facilitate generalization over the label distribution.

\end{document}